\documentclass[a4paper,11pt,twocolumn,final]{article}

\usepackage{mathptmx}
\usepackage{icphs2015}

\usepackage{setspace}
\usepackage{microtype}
\titlespacing*{\section}{0pt}{9pt}{9pt}
\titlespacing*{\subsection}{0pt}{8pt}{8pt}
\titlespacing*{\subsubsection}{0pt}{8pt}{8pt}

\usepackage{tipa}

\usepackage[para]{footmisc}
\renewcommand\footnotemark{}

\usepackage[inline, shortlabels]{enumitem}

\usepackage[numbers]{natbib}

\usepackage[acronym, shortcuts]{glossaries}
\newacronym{ema}{EMA}{electromagnetic articulography}
\newacronym{icp}{ICP}{iterative closest point}
\newacronym{mri}{MRI}{magnetic resonance imaging}
\newacronym{pca}{PCA}{principal component analysis}
\newacronym{tps}{TPS}{thin plate spline}
\newacronym{uti}{UTI}{ultrasound tongue imaging}

\usepackage[hidelinks, pdfusetitle]{hyperref}

\usepackage{amssymb}

\usepackage{siunitx}
\title{A statistical shape space model of the palate surface trained on 3D MRI scans of the vocal tract}
\author{%
  Alexander Hewer\textsuperscript{1--3},
  Ingmar Steiner\textsuperscript{2,3},
  Timo Bolkart\textsuperscript{1,3},
  Stefanie Wuhrer\textsuperscript{4},
  Korin Richmond\textsuperscript{5}
}
\organization{%
  \vspace{-\baselineskip}
  \singlespacing
  \textsuperscript{1}Saarbrücken Graduate School of Computer Science, Germany\\
  \textsuperscript{2}DFKI Language Technology Lab, Saarbrücken, Germany\\
  \textsuperscript{3}Cluster of Excellence Multimodal Computing and Interaction, Saarland University, Germany\\
  \textsuperscript{4}INRIA Grenoble Rhône-Alpes, France\\
  \textsuperscript{5}Centre for Speech Technology Research, University of Edinburgh, UK\\
  ~\thanks{%
    This study uses data from work supported by EPSRC Healthcare Partnerships Grant number EP/I027696/1 (``Ultrax'').
  }
}
\email{%
  \vspace{-\baselineskip}
  \{ahewer|steiner|tbolkart\}@mmci.uni-saarland.de,
  stefanie.wuhrer@inria.fr,
  korin@cstr.ed.ac.uk
}
\newcommand*{\ourkeywords}{vocal tract MRI, principal component analysis, palate model}
\hypersetup{%
  pdfinfo={%
    Author={Alexander Hewer, Ingmar Steiner, Timo Bolkart, Stefanie Wuhrer, Korin Richmond},
    Keywords={\ourkeywords},
    Subject={ICPhS 2015}
  }
}

\begin{document}

\maketitle

\begin{abstract}
We describe a minimally-supervised method for computing a statistical shape space model of the palate surface.
The model is created from a corpus of volumetric \ac{mri} scans collected from 12 speakers.
We extract a 3D mesh of the palate from each speaker, then train the model using \ac{pca}.
The palate model is then tested using 3D \ac{mri} from another corpus and evaluated using a high-resolution optical scan.
We find that the error is low even when only a handful of measured coordinates are available.
In both cases, our approach yields promising results.
It can be applied to extract the palate shape from \ac{mri} data, and could be useful to other analysis modalities, such as \ac{ema} and \ac{uti}.
\end{abstract}

\glsresetall

\section{Introduction}

The palate plays an important role in articulation;
as part of the vocal tract walls it contributes to vowel production, and it is critical for the production of obstruents such as /\textyogh/, /\textesh/, or /j/, and for palatalization \cite{Hiki1986}.
Therefore, analyzing its shape and understanding its interaction with other articulators is of great interest in speech science.
A shape model of the palate could also contribute to acoustic models of the vocal tract.

Direct measurements of the palate shape are however a challenging task.
Nowadays, \ac{mri} is the modality of choice for imaging the human vocal tract.
This technique is able to provide dense 3D information about the inside of a speaker's mouth without being hazardous or invasive.
The acquired data, however, has to be further processed to obtain the desired palate shape.
In particular, a high-level structured shape representation is desirable, such as a polygonal mesh.

A model of the palate surface can be directly used in various fields of application.
For example, for automatic image segmentation of \ac{mri} data, it can be used as a prior.
It could also provide a persistent landmark for analysis with spatially sparse modalities, such as \ac{ema} or \ac{uti}.
Moreover, a palate mesh could be integrated to derive the vocal tract area function for acoustic modeling.

\subsection{Related work}

Analyzing the shape of the palate is an active field of research.

Yunusova et al.\ \cite{Yunusova2012} used a \ac{tps} technique to estimate the contour of the palate in a palate trace acquired by \ac{ema}.
\Ac{tps} is a data-driven method that tries to deform a thin plate such that it passes through a set of control points.
Additionally, the resulting plate should have some degree of smoothness.
In their work, Yunusova et al.\ found that the weight for the smoothness constraint had an impact on the result:
using values that were too small resulted in an overfitting to the sample points of the palate trace, whereas too large a value prevented the plate from deforming at all.
In their experiments, they derived an optimal value for this weight empirically.
As the method is purely data-driven, it might produce undesirable results if the data is too sparse.

Lammert et al.\ \cite{Lammert2013} used realtime \ac{mri} to investigate the morphological variation of the palate and the posterior pharyngeal wall.
They extracted the shape information from mid-sagittal slices of the vocal tract.
Afterwards, they applied a \ac{pca} to the obtained data to extract the principal modes of variation of both structures.
In their study, they found that the obtained principal modes could actually be related to anatomical variation, such as the degree of concavity of the palate.
However, this study was restricted to the 2D case.

\subsection{Our contribution}

In this work, we present a minimally-supervised method for training a statistical model of the 3D shape of the palate surface.
Our approach consists of two steps.
We first extract the full shape of the palate surface from static 3D \ac{mri} scans of different speakers, where we use a polygonal mesh as the shape representation.
Afterwards, we apply a \ac{pca} to this data in order to train the model.
As the whole process is minimally-supervised, it is relatively easy to include additional \ac{mri} scans to improve the coverage of the model.

Such a statistical model can be helpful:
for example, it could be useful for investigating the anatomical variation of the 3D shape of the human palate.
Moreover, it represents a shape space that is able to generate new palate shapes and evaluate the probability of a specific shape.
This property can be used for detecting and reconstructing the shape of the palate form data that is incomplete or very sparse, such as \ac{ema} or \ac{uti}.

\section{Methods}

\begin{figure}
  \caption{%
    Palate mesh with landmark vertices shown as colored spheres.
    {\bfseries Left:} View from the top.
    {\bfseries Right:} Side view.
    \label{fig:meshlandmarks}
  }
  \centering
  \includegraphics[width=\linewidth]{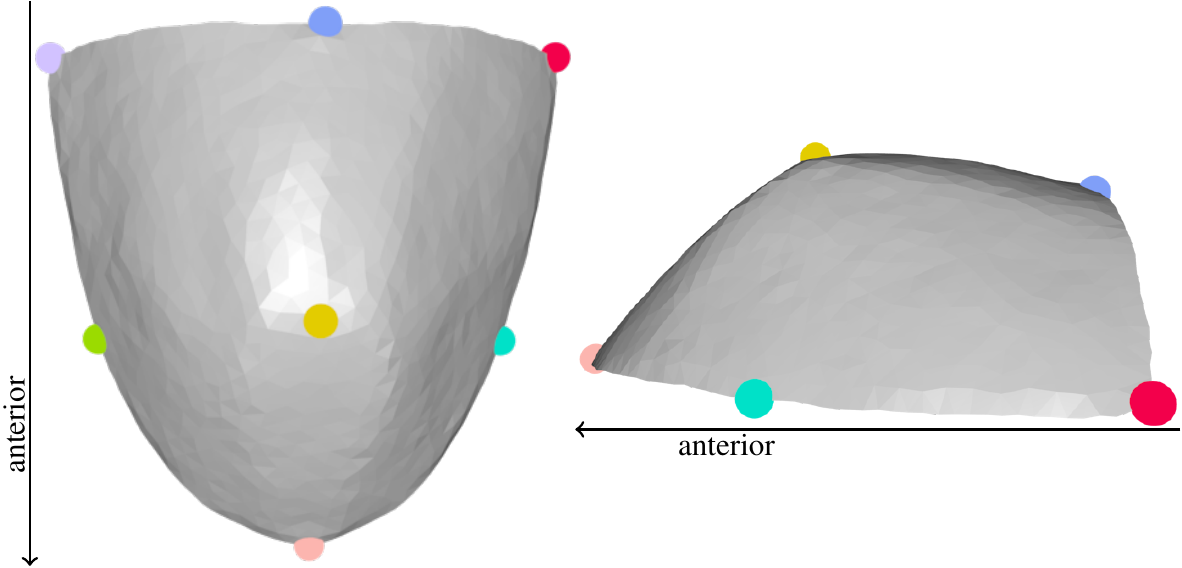}
\end{figure}

Before outlining our approach, we first want to give a definition of the polygonal mesh \(M := ( V, F ) \) used as a shape representation.
\(V := \{\vec{v_i}\}\) with \(\vec{v_i} \in \mathbb{R}^3\) is called the vertex set of the mesh and \(F\) its face set. 
A face \( f \in F \) is a set of vertices that form a surface patch in the form of a polygon, e.g., a triangle, if linked by edges.
Stitching all faces together results in the full surface.
An example mesh can be seen in \autoref{fig:meshlandmarks}.

\subsection{Shape extraction}
\label{sec:shape-extraction}

In the first step of our approach, we focus on extracting 3D meshes representing the palate surface from volumetric vocal tract \ac{mri} scans collected from a number of different speakers (cf.\ \autoref{sec:mri-data}, below).
Here, we are using the minimally-supervised method of \cite{Hewer2014} that can be summarized as follows:

\subsubsection{Surface point extraction}

First, the region belonging to tissue is identified using an automatic image segmentation technique.
In particular, each scan is interpreted as a 3D image with gray values in the interval \([0, 255]\).
In our case, we chose to use a basic thresholding method to identify the tissue:
some tissue like the palate surface appears much brighter than material with lower hydrogen density, such as air or bone.
Thus, we automatically classify each point with a brightness higher than some threshold value as tissue.

The method then proceeds by extracting the surface points of the identified tissue regions, which produces a point cloud \( P := \{\vec{p_i} \} \) with \(\vec{p_i} \in \mathbb{R}^3\).

\subsubsection{Template fitting}

Then, a template fitting technique is applied to align a provided template mesh to the obtained point cloud.
Two manual components were required for this step, viz.\
\begin{enumerate*}[\itshape (a)]
  \item a template mesh for the palate surface created beforehand from a single 3D \ac{mri} scan, using a medical imaging software \cite{Rosset2004}; and
  \item a set of 7 manually selected vertices used as landmarks in a rigid alignment initialization step (cf.\ \autoref{sec:rigid-alignment}).
\end{enumerate*}
These are required to identify the correct subset of points representing the palate and to deform the template mesh accordingly.

The palate mesh extraction step results in a collection of training meshes \(M_i = ( V_i, F ) \) with \( i \in [1, n] \), where $n$ is the number of speakers.
We remark that the meshes may differ in the position of their vertices, i.e., \(\vec{v_k} \in V_i \neq \vec{v_k} \in V_j\) for \(i \neq j\).
Their faces, however, still consist of the same vertices which differ only in their position.

\subsection{Training the model}

\begin{figure}
  \caption{%
    Landmarks selected on an \ac{mri} scan.
    {\bfseries Left:} Sagittal slice.
    {\bfseries Right:} Coronal slice.
    \label{fig:user-landmarks}
  }
  \centering
  \includegraphics[height=0.5\linewidth]{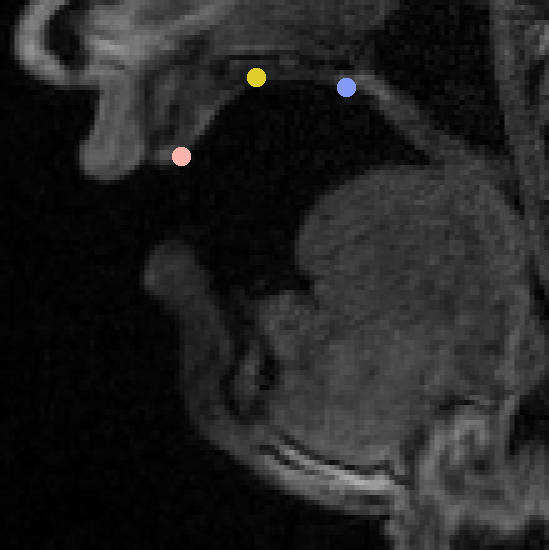}
  \hspace{10pt}
  \includegraphics[height=0.5\linewidth]{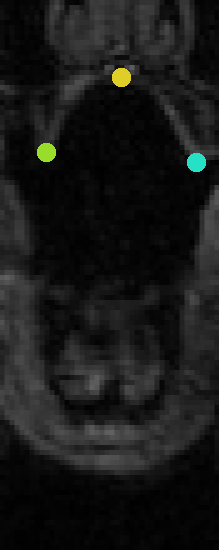}
\end{figure}

In order to train our statistical model, we have to ensure that the palate meshes only differ from each other in their shape.
To this end, we first apply a Procrustes alignment \cite{Dryden1998} to the collection of all extracted meshes.
This serves to remove differences in their location, orientation, and scale, which enables us to analyze the features of their shape.

Next, we convert the transformed meshes to feature vectors such that the coordinates belonging to each vertex are located in consecutive rows.
Finally, we apply a \ac{pca} to these vectors.
Such methods are often used in literature to analyze data, cf., e.g., \cite{Allen2003, Brunton2014, Cootes2001, Cootes1995}.
This provides us with the set of principal directions \(\vec{e_i} \in \mathbb{R}^k\) with \(k = 3 |V|\) of the training data.
Interpreting these vectors as a basis gives us access to a space of palate shapes.
Afterwards, we project the training data into this shape space and learn its probability distribution by fitting a multivariate Gaussian \cite{Davies2008}.
Thereby we obtain the variances \(\lambda_i \in \mathbb{R}\) and means \(m_i \in \mathbb{R}\) along the associated principal directions \(\vec{e_i}\) of our training data.
Thus, we can also measure the probability of a specific shape in our learned shape space.

\subsection{Generating palate shapes}

The trained model can be used as follows to generate a new palate mesh \(M^* = (V^*, F^*) \):
first, we generate a vector \(\vec{x}\) representing a palate shape by computing
\begin{equation}
  \vec{x} = \sum_i \big( ( m_i + c_i ) \vec{e_i} \big) = \vec{m} + \sum_i \left( c_i \vec{e_i} \right)
\end{equation}
where \(\vec{m}\) is the mean of our training data and \( c_i \in \mathbb{R} \) is the provided coefficient for the principal direction \(\vec{e_i}\).
Then, we convert \(\vec{x}\) to a vertex set \(V^*\) and assign \(F^* = F\), the face set of our template mesh.

\subsection{Using the model to register new data}

In order to use our model to reconstruct a palate from a point cloud, we perform the following steps:

\subsubsection{Rigid alignment}
\label{sec:rigid-alignment}

First, we have to find the optimal scale and location in the point cloud for the mesh generated by the mean of our model.
This step is necessary because our shape space is not able to produce rigid transformations like translations or rotations.
We use the following approach to facilitate this process:
on the mesh, we selected 7 vertices as landmarks, as shown in \autoref{fig:meshlandmarks}.
Here, we see that we used three landmarks along the mid-sagittal line of the palate:
one at the incisors, one at the hard/soft palate boundary, and another at the point of greatest curvature.
In order to add lateral information, we used the latter two landmarks as the anchor for two additional landmarks at either side of the palate.

Afterwards, we find the points in the data corresponding to these landmarks.
If the used cloud originates from an \ac{mri} scan, this scan can be used to derive the coordinates like in \autoref{fig:user-landmarks}.
Here, it is evident that the landmark locations are relatively easy to identify for a user.
The scale and position of the mesh are then determined by finding the best rigid transformation that maps the user-provided coordinates to the landmarks on the mesh.
Additionally, an \ac{icp} approach \cite{Besl1992} was applied to further improve this rigid alignment.

\subsubsection{Fitting the model}

In the final step, we find the coefficients \(c_i\) for the principal directions of our model such that the resulting mesh is near the data in the provided point cloud.
However, we limit the values for the coefficient \(c_i\) to the interval \([- \sqrt{\lambda_i}, \sqrt{\lambda_i}]\).
This means we only consider values with a distance of no more than \(\sqrt{\lambda_i}\) from the corresponding mean \(m_i\) of the coefficient in the training data, which serves to avoid unlikely palate shapes and prevent overfitting.
In order to find these coefficients, we minimize an energy where we use a quasi-Newton scheme \cite{Liu1989} to find a minimizer.

We note that this approach to fitting the model to the data is more robust than applying a template fitting technique directly.
The model contains a whole space of palate shapes, whereas a template only represents a single shape.
In contrast to a template mesh, it also allows to evaluate the probability of a generated shape.
Furthermore, a template fitting offers many more degrees of freedom to align the template to the data, which means that it would also be possible for a palate mesh to be deformed into an implausible shape.

\section{Datasets}
\label{sec:mri-data}

We used scans from two datasets for training our model:
the full dataset of the Ultrax project \cite{Ultrax2014} and that of Adam Baker \cite{Baker2011}.
Both were recorded using a Siemens MAGNETOM Verio at the Clinical Research Imaging Centre in Edinburgh for the purpose of observing the vocal tract configuration for different phones.
The Baker dataset consists of static 3D \ac{mri} scans of a single male speaker.
It was recorded as part of the Ultrax project, but released separately.
The Ultrax dataset itself contains static 3D \ac{mri} scans of 11 adult speakers where seven are female and four are male.
Each considered scan consists of 44 sagittal slices with a thickness of \SI{1.2}{mm} and size (whole head) of \num{320 x 240} pixels with a voxel size of \SI{1.1875 x 1.1875 x 1.2}{\mm}.

To evaluate our approach, we used the volumetric \ac{mri} subset of the mngu0 corpus \cite{Steiner2012}, which contains data from one male speaker, including high-resolution 3D scans of a plaster cast of his teeth and palate.
Here, each \ac{mri} scan consists of 26 sagittal slices of \SI{4}{mm} thickness.
The size of each slice is given by \num{256 x 256} pixels with a corresponding voxel size of \SI{1.1 x 1.1 x 4}{\mm}.
We see that compared to the Ultrax data, these scans offer a lower spatial resolution along the sagittal dimension.

\section{Experiments}

\begin{figure}
  \caption{%
    Colored fitted palate mesh of first experiment and maxilar dental cast.
    Color indicates distance to nearest point on dental cast.
    \label{fig:mesh-on-dentalcast}
  }
  \centering
  \includegraphics[width=.9\linewidth]{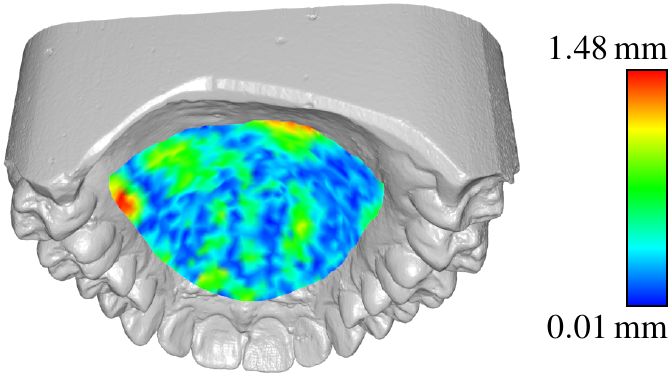}
\end{figure}

For the training, we used all twelve speakers of the Baker and Ultrax datasets.
We selected for each speaker a scan where the palate was clearly visible, with no lingual contact.
We then cropped each scan to a region of interest containing only the palate in order to reduce the memory requirements for the point cloud.
Afterwards, we applied the methods described in \autoref{sec:shape-extraction} to extract the palate shapes.
Here, we used the value \(t = 25\) for the thresholding parameter to perform the image segmentation.
All extracted meshes were then used to train the model.

\subsection{Experiment setup}

In the first experiment, we wanted to investigate if our model could handle data of a speaker it was not trained with.
To this end, we selected the /\textturnscripta/ scan of the volumetric data of mngu0.
We prepared the data as follows:
the scan was once again cropped to a region containing only the palate.
We then extracted the surface points of the tissue where we used the threshold \(t = 25\) in the image segmentation step.
Additionally, we distributed the landmarks needed for the rigid alignment by using the cropped scan.
Finally, we fitted our trained model to the obtained point cloud.

Afterwards, we analyzed in a second experiment how our model behaves if \emph{only} the 7 landmarks chosen in the first experiment are used for the fitting.

\subsection{Evaluation}

\begin{figure}
  \caption{%
    Cumulative error functions for the two experiments.
    \label{fig:cdfs}
  }
  \centering
  \includegraphics{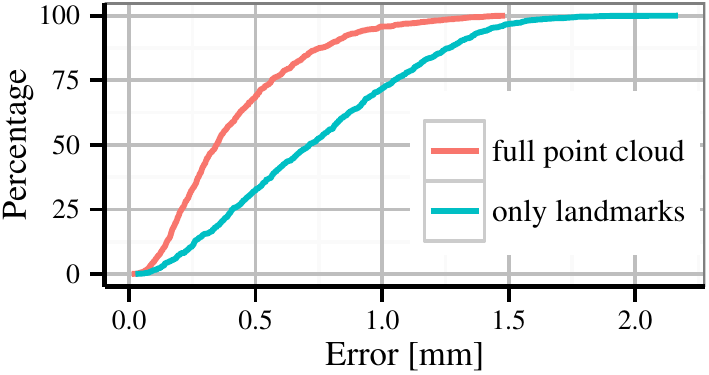}
\end{figure}

We used the maxillary dental cast of the speaker as the reference solution for the shape of the hard palate.
However, the obtained palate meshes differed from the dental cast in their location and orientation.
Therefore, we again used landmarks and an \ac{icp} technique to perform a rigid alignment to remove these differences. 
This time, no scaling was applied in order to preserve the original shape.
We then measured for each vertex of the palate mesh the distance to the closest point on the dental cast.
A heat map visualizing these distances for the mesh of the first experiment can be seen in \autoref{fig:mesh-on-dentalcast}.
Afterwards, we interpreted this distance as an error measure and computed the cumulative error function.
In \autoref{fig:cdfs}, we see that in the first experiment nearly \SI{75}{\percent} of the error are below \SI{0.5}{mm}.

For the second experiment using only the seven landmark points, nearly \SI{75}{\percent} of the error are below \SI{1}{mm}, which indicates that our model can produce acceptable results even with only very sparse information.

\section{Conclusion}

In this work, we described a minimally-supervised method for training a statistical shape model of the palate surface.
We saw that a model trained with the palate shapes of twelve speakers was already useful.
In particular, we found that it could be used to extract palate information from an \ac{mri} scan of a new speaker.
Furthermore, even when only a handful of points are used, our model can be fit with acceptable precision, which allows us to use sparse input data, such as from an \ac{ema} palate trace.

Further experiments are scheduled to obtain reference data from more speakers, using an intraoral scanner, such as a 3shape TRIOS.
Moreover, we plan to investigate how the trained model can be used to reconstruct palate information from existing \ac{ema} data of these speakers.
Moreover, we plan to acquire more \ac{mri} data to increase our training set.

\clearpage
\bibliographystyle{icphs2015}
\bibliography{references}

\end{document}